\documentclass[
twocolumn,
]{ceurart}

\usepackage{listings}
\usepackage{adjustbox}
\usepackage{paralist}
\usepackage[latin1]{inputenc}
\usepackage{amssymb,amsmath,array}
\usepackage{booktabs}
\usepackage{svg}
\usepackage{caption}
\usepackage{subcaption}

\begin{document}

\copyrightyear{2022}
\copyrightclause{Copyright for this paper by its authors.
  Use permitted under Creative Commons License Attribution 4.0
  International (CC BY 4.0).}

\conference{1st International Workshop on Pervasive Artificial Intelligence, Hosted by the 2022 IEEE World Congress on Computational Intelligence}

\title{Continual-Learning-as-a-Service (CLaaS): On-Demand Efficient Adaptation of Predictive Models}

\author[1]{Rudy Semola}[%
email=r.semola@phd.unipi.it,
url=https://github.com/rudysemola,
]
\cormark[1]
\address[1]{Department of Computer Science, University of Pisa}

\author[1]{Vincenzo Lomonaco}[%
email=vincenzo.lomonaco@unipi.it,
url=https://www.vincenzolomonaco.com,
]

\author[1]{Davide Bacciu}[%
email=davide.bacciu@unipi.it,
url=http://pages.di.unipi.it/bacciu,
]

\cortext[1]{Corresponding author.}

\maketitle

\begin{abstract}
    Predictive machine learning models nowadays are often updated in a stateless and expensive way.
    The two main future trends for companies that want to build machine learning based applications and systems are \emph{real-time inference} and \emph{continual updating}.
    Unfortunately, both trends require a mature infrastructure that is hard and costly to realize on-premise.
    This paper defines a novel software service and model delivery infrastructure termed Continual Learning-as-a-Service (CLaaS) to address these issues. 
    Specifically, it embraces continual machine learning and continuous integration techniques. 
    It provides support for model updating and validation tools for data scientists without an on-premise solution and in an efficient, stateful and easy-to-use manner. 
    Finally, this CL model service is easy to encapsulate in any machine learning infrastructure or cloud system.
    This paper presents the design and implementation of a CLaaS instantiation, called Continual Brain, evaluated in two real-world scenarios. The former is a robotic object recognition setting using the CORe50 dataset while the latter is a named category and attribute prediction using the DeepFashion-C dataset in the fashion domain. Our preliminary results suggest the usability and efficiency of the Continual Learning model services and the effectiveness of the solution in addressing real-world use-cases regardless of where the computation happens in the continuum Edge-Cloud.
\end{abstract}


\section{Introduction}
\label{sec:intro}

Machine Learning (ML) has become a fast-growing, trending approach in solution development in practical scenario.
Deep Learning (DL) which is a subset of ML, learns using deep neural networks to simulate the human brain. 
Nowadays, the latter is widely adopted in a variety of applications, especially in Computer Vision (CV) \cite{Alzubaidi2021ReviewOD} \cite{Zou2019ObjectDI} and Natural Language Processing (NLP) \cite{SurvayNLP} \cite{AttentionNLP}.
Both academia and industrial research are investing significant efforts in developing ML-as-a-Service (MLaaS) tools to build and monitor a variety of smart applications \cite{uberMLaaS} \cite{MLaaS2015}.

Current technical issues related to software development and delivery in organizations that work on ML projects have brought novel practicals and concepts.
In particular, the integration of Machine Learning practices that support data engineering, with the Development and Operations (DevOps) practices based software development, has resulted in Machine Learning Model Operationalization Management (MLOps) \cite{MLOps}.
MLOps principles have been proposed and used to deploy and maintain machine learning models in production reliably and efficiently. 
It incorporates ML models for solution development and maintenance with continuous integration (CI) to provide efficient and reliable service. 
Different roles such as data scientists, DevOps engineers, and IT professionals are involved in different MLOps process.

The accuracy of the predictions made by ML applications depends on many factors such as data type, training algorithm, hyperparameters, learning rate and optimizers. 
Different edge or cloud based applications need the latest real-time data and are retrained frequently to produce more accurate and precise predictions. 
Thus, the training models should be retrained without human intervention using reproducible and automating pipelines in a continuous manner. 
It is challenging to automate these decisions making processes using current MLOps. 
Moreover, these MLOps toolkits should be user-friendly, reliable, and efficient to use in industrial and practical scenarios.

Currently, we note also that ML serving systems are not able to handle the dynamic and non-stationary production environments adequately.
The principal causes come from the concept drift \cite{DeepMindOvercoming} issue of real-life data.
One of the main consequences is decay in the performance of the model. The second is monitoring data to understand when a drift distribution happens.
To overcome these issues, we note that the companies innovative trend is toward real-time inferences and Continual Learning (CL) update models  \cite{TowardCL}. 
The first to generate more accurate predictions, and the latter to adapt models to changing non-stationary production environments \cite{CLRobotics}.

To support the generation of CL solutions for research and practical applications, several systems have been developed, as summarized in Table \ref{tab:CL-framework}.
However, most of these solutions do not support CL strategies for continuous training (CT).
Several CL tools are widely used by research communities, but few of them meet the requirements of real industrial scenarios.

\subsection{Proposal Solution and Scientific Value}
In this paper, we propose a new model service paradigm named Continual-Learning-as-a-Service (CLaaS) with the following guidelines.
First, CLaaS should be easily integrated with existing ML serving systems and MLOps toolkit either on-primes edge or cloud-based.
Second, we are interested not only for research purposes but also in fast R\&D prototype projects. Therefore, it allows for training and updating ML models in existing serving systems in an efficient, scalable and adaptable manner.  
Third, this model service can allow the building and maintaining of low-cost smart infrastructure without particular knowledge of Continual Learning.

We start to define CLaaS paradigm and compare it to existing related continual ML model services (Section \ref{sec:claas-parag}). 
Then, the architecture and system implementation details of a possible CLaaS instance we called \emph{Continual Brain} are described (Section \ref{sec:system-DI}). Some preliminary studies, comparing the Cumulative (when a new batch of data becomes available, prediction models are re-trained from scratch) \cite{CLRobotics} approach and well-known CL strategies, are also executed in two particular practical domains to demonstrate the potential advantages in terms of efficiency (Section \ref{sec:demo}).  
To sum up, future improvements and optimization directions are presented (Section \ref{sec:conlusion}). 

\begin{table*}[hbt!]
\caption{Comparison of several (continual) machine learning operations toolkits. HPO: hyper parameter optimization in training stage; MLOps: Model Operationalization Management; CM: Continuous Monitoring; CT: Continuous Training.}
\centering
\begin{adjustbox}{width=1\textwidth}
\small
\begin{tabular}{llllllll} 
\toprule
    \textbf{Tools}  & \textbf{Drift data} & \textbf{CL zoo} & \textbf{CL} & \textbf{CM} & \textbf{CT } & \textbf{Portability} & \textbf{auto} \\
     	&  \textbf{detection} & \textbf{algorithm} & \textbf{metrics} & \textbf{(MLOps)} & \textbf{(MLOps)} & \textbf{(edge-cloud)} & \textbf{HPO} \\
\midrule
Continual \cite{CLinPracticeAWS}	& \checkmark & - & - & \checkmark & \checkmark & - & \checkmark \\
AutoML (AWS)	&  &  &  &  &  &  & \\
\midrule
AWS Sagemaker 	& \checkmark & - & - & \checkmark & \checkmark & \checkmark & \checkmark \\
\cite{AWSsagemaker}	&  &  &  &  &  &  & \\
\midrule
Avalanche \cite{lomonaco2021avalanche}	& - & \checkmark  & \checkmark  & - & -  & - & - \\
\midrule
ModelCI-e \cite{huang2021modelci}	& \checkmark & \checkmark & - & \checkmark & \checkmark & \checkmark & - \\
\midrule
CLaaS 	& \checkmark & \checkmark & \checkmark & \checkmark & \checkmark & \checkmark & \checkmark \\
\midrule
\textbf{Continual Brain} 	& - & \checkmark  & \checkmark & - & \checkmark  & \checkmark & - \\
\bottomrule
\end{tabular}
\end{adjustbox}
\label{tab:CL-framework}
\end{table*}

\section{Related Work and Background}
\label{sec:related-work} 
This section first summarizes MLOps methodologies with a brief overview of the tools support. 
Then, it describes Continual Learning and its effectiveness in real-world contexts.

\subsection{Machine Learning Operations (MLOps) and Tool support}
MLOps is a set of scientific principles, tools, and techniques of Machine Learning and traditional Software Engineering to design and build complex computing systems. 
It encompasses all stages from data collection, to model building, to making the model in SW production system.
MLOps emerges from the understanding that separating the ML model development from the process that delivers it, named ML operations, lowers the quality, transparency, and agility of the whole intelligent software \cite{MLOps}.
For More detail, challenge and commercially available MLOps tool support in software development are well described in \cite{MLOpsTools} and \cite{MLOps2}.

\begin{figure*}[ht!]
  \centering
   \includegraphics[width=0.98\linewidth]{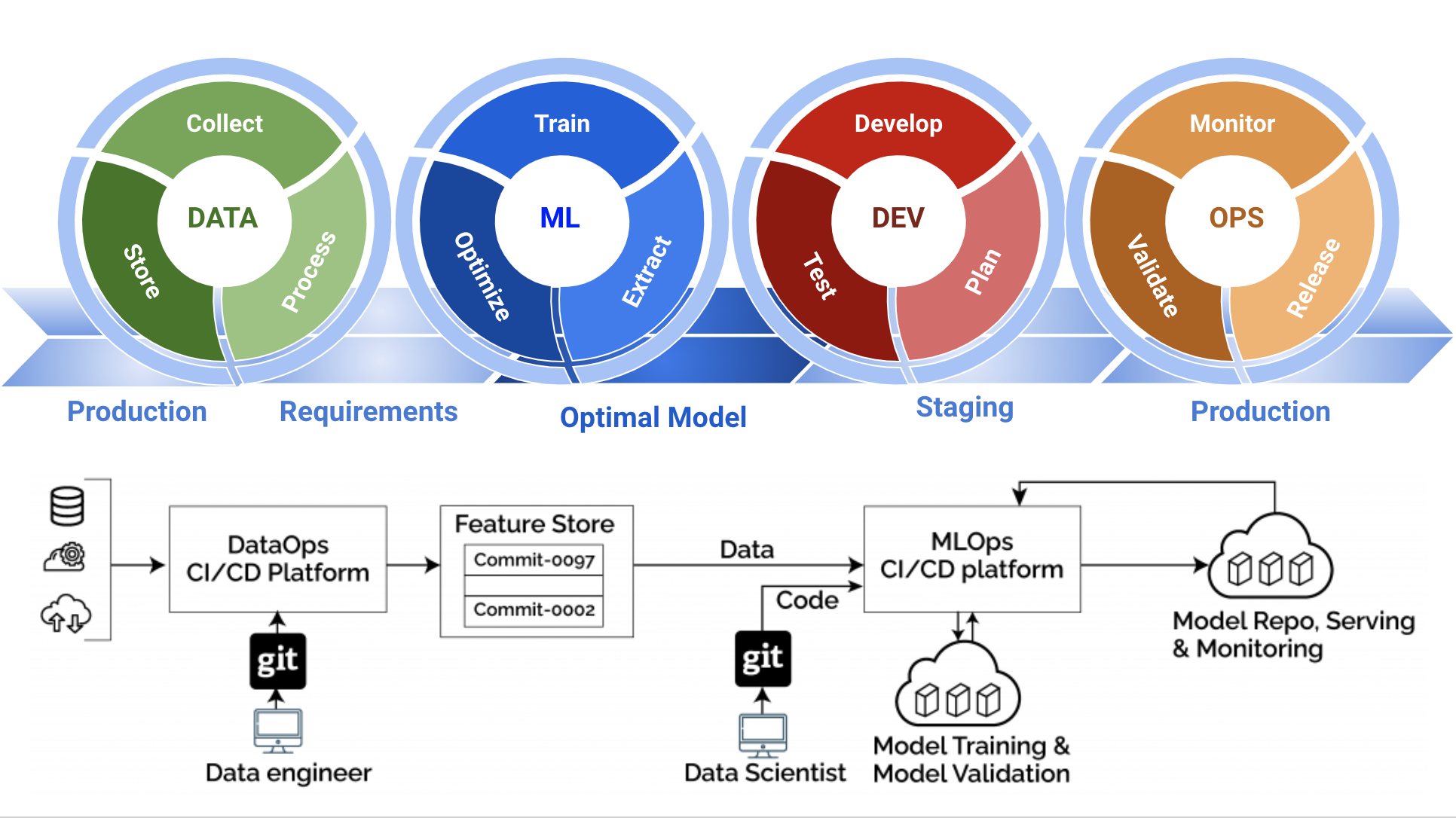}
   \caption{Graphical representation of  MLOps, an end-to-end Machine Learning life-cycle management. It is an Iterative-Incremental Process mainly based on 3 stages: Data, Machine Learning (ML) and DevOps. The latter has two principles named Continuous Integration (CI) and Continuous Delivery (CD).
 MLOps adds other two practices named Continuous Monitoring (CM) and Continuous Training (CT).
   This work is focused on the Machine Learning development stage to train and validate the models with MLOps practices and Continual Learning methodologies.}
   \label{fig:MLOps1}
\end{figure*}

Typical workflow for machine learning-based software development includes three primary subjects. They are data, ML model, and code and this workflow consists of three main phases 
\begin{itemize}
    \item \emph{Data Engineering}, concerned with data acquisition and data preparation
    \item \emph{ML Model Engineering}, in which the process starts from model training, evaluation and serving
    \item \emph{Code Engineering}, which it is integrated the ML model into the final product
\end{itemize}

There are three main problems that influence the value of ML models once they are in production.
First, ML models are sensitive to the semantics, amount and completeness of incoming data (data quality).
Second is the performance degradation of ML models in production over time (model decay). In fact, real-life data are non-stationary and most of the time they have not been seen during the model training. 
Third, when transferring ML models to new business customers, these models, which have been pre-trained on different user demographics, might not work correctly according to quality metrics (locality).

MLOps try to establish effective practices and processes around designing, building, and deploying ML models into production.
Currently, MLOps tools range from using a machine-learning platform to implementing an on-premise solution by composing open-source libraries.

MLOps platforms should support languages, frameworks and libraries in an easy-to-use unique environment.
Although cloud service providers have similar platforms, they are costly and are not addressing the ML problem itself through a single dashboard. 
In addition to that, some of the platforms do not offer free licenses to use as embedded systems.
The accuracy of the predictions made by DL models depends on many factors and some applications need the latest real-time data that are retrained frequently to produce more accurate and precise predictions. 
Thus, the training models should be retrained without human intervention using reproducible pipelines and in a continuous manner.
It is challenging to automate these decisions making processes using the current MLOps tools. 
Continuous training and evaluation techniques and strategies have to take into account the non-stationarity nature of the data. 
Continual Learning methodologies could be beneficial for these MLOps tools. 

\begin{figure*}[ht!]
  \centering
   \includegraphics[width=0.98\linewidth]{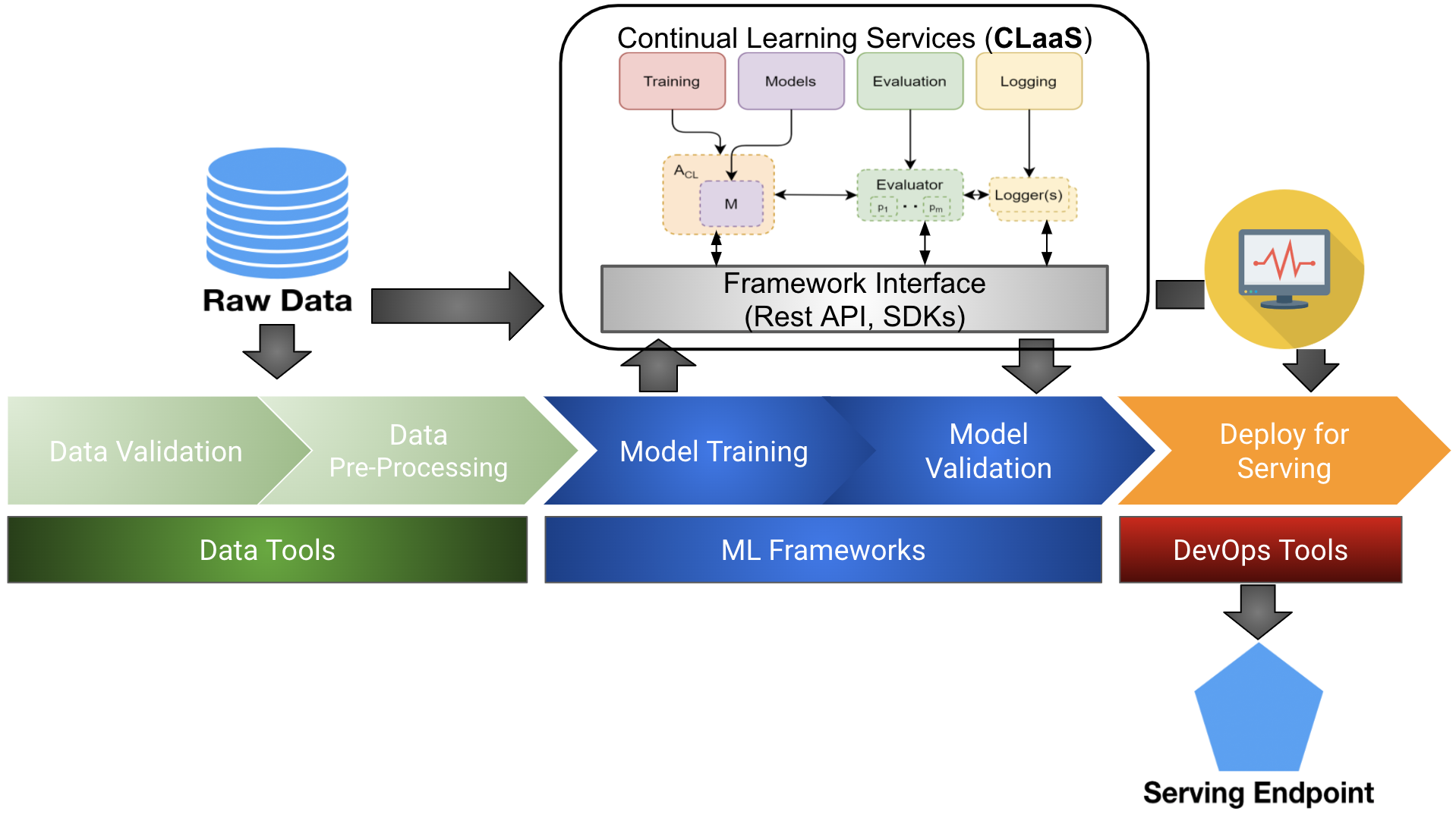}
   \caption{Graphical representation of a typical Machine Learning pipeline. This approach includes three procedures: $(1)$ collection, selection and preparation of data to be used in model training, $(2)$ finding and selecting the most efficient model after validation and $(3)$ sending the selected model to the serving system. CLaaS is a service model approach that helps data scientists and engineers to monitor and update continually and efficiently the models over time.}
   \label{fig:MLpipeline}
\end{figure*}

\subsection{Continual Learning and Real-World Applications}

In a classical Continual learning problem,  a single neural network model needs to sequentially learn a series of tasks. 
During training, only data from the current task is available and the tasks are assumed to be clearly separated. 
Continual learning methodologies deal with ML problems in a non-stationary data setting. 
They try to train models in a sequence of tasks to acquire new knowledge without forgetting what has been trained in the past. 
This problem has been actively studied in recent years and many methods for alleviating catastrophic forgetting have been proposed \cite{3CLscenario}.
Unlike closed or simulation environments the data does not follow a stationary distribution in real applications. 
In a real-world application, we have a constant flow of information, where the distribution can change due to various external or internal factors. 
This problem creates the need to update the model continually efficient and adaptively manner.

Recent and relevant continual learning studies for different applications have been done. 
From surveillance videos, robotics and machine vision, clinical and medical sector, to the industry with edge and cloud computing \cite{hayes2022online} \cite{Doshi_2020_CVPR_Workshops} \cite{kiyasseh2020clops}.
These works demonstrate the usability and the effectiveness of CL in practical several domains.

At the same time, there are also many real-world and application-research directions unexplored with Continual Learning. 
Currently, companies interested to explore CL for their business purpose are dealing with costly and quite challenging approaches.
It is required a research team and toolkit for expert of the CL domain.
Furthermore, enabling CL in production is still a challenging problem. 
These problems motivated the design of a new service paradigm that embrace Continual learning Methodologies and tool in a cheaper and easy-to-use manner for real-world scenarios.

\subsection{Avalanche, End-to-End Library for Continual Learning}
In \cite{lomonaco2021avalanche} it is proposed Avalanche, an open-source end-to-end library for continual learning research based on PyTorch. Avalanche provides a shared and collaborative codebase for fast prototyping, training, and reproducible evaluation of continual learning algorithms \cite{lomonaco2021avalanche}.

Avalanche is designed with five main principles important to CL research and real-world applications. 
These principles are  \emph{Comprehensiveness and Consistency};\emph{Ease-of-Use}; \emph{Reproducibility and Portability}; \emph{Modularity and Independence}; \emph{Efficiency and Scalability}.
At the current stage of development, the library is organized into five main modules: \emph{Benchmarks}, \emph{Training}, \emph{Evaluation}, \emph{Models} and \emph{Logging}.

Avalanche implements a system of \emph{Plugins} to facilitate the customization of strategies, metrics and logging.
This is used by strategies, metrics, and loggers. 
It allows them to interact with the training loop and execute their code at the correct points using a simple interface.
In particular, \emph{SupervisedPlugin} is the base class for plugins for a supervised scenario, from which loggers inherit, and PluginMetric, a base class for metrics.

\section{Continual-Learning-as-a-Service Paradigm} 
\label{sec:claas-parag} 
\begin{figure*}[ht!]
  \centering
   \includegraphics[width=0.975\linewidth]{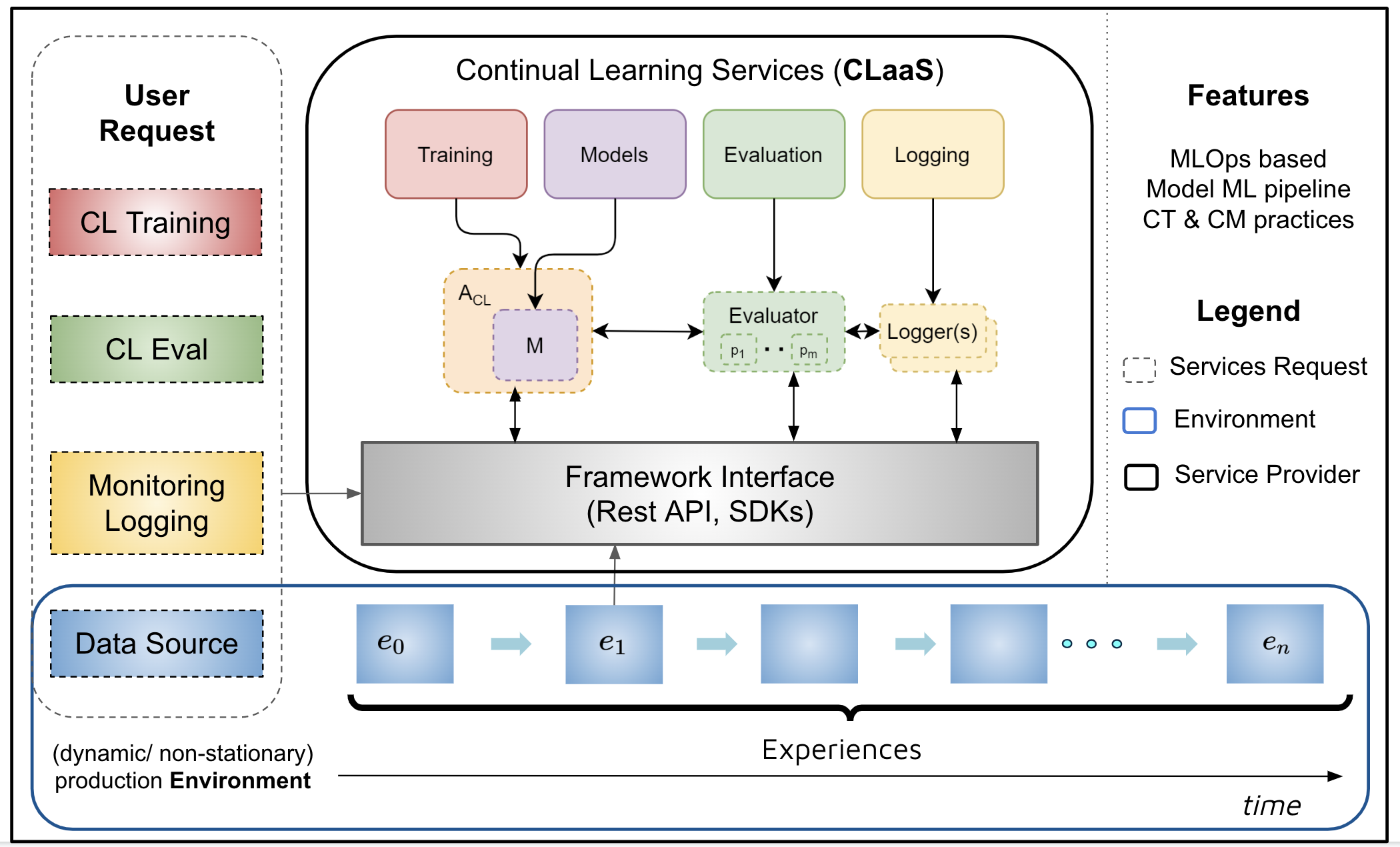}
   \caption{Graphical representation of the CLaaS paradigm. It is mainly based on MLOps practices and Continual Learning methodologies.}
   \label{fig:claas-parag}
\end{figure*}

Continual-Learning-as-a-Service is a service model paradigm mainly based on Continual Learning methodologies to continuously monitor data distribution shifts and update the model in an efficient fashion.
As presented in Figure \ref{fig:claas-parag}, CLaaS are based on MLOps principles. In particular, Continuous Training (CT) and Continuous Monitoring (CM) are Continual Learning based. It is easy to use in an on-demand manner for both edge and cloud-based systems.
This Section start from a comparative description of (continual) machine learning operations toolkits and follow with the motivations and benefits of the use of the CL features in this paradigm


\subsection{Comparison of (continual) machine learning operations toolkits}
The main features and differences between the other (continual) Machine Learning tools for MLOps are summarized in Table \ref{tab:CL-framework}.  
Most of monitoring solutions are focused on analyzing statistics of a feature and alert when significant changes in these statistics happen. 
The level of automation of the ML pipeline defines the maturity of the ML process, which reflects the velocity of training new models given new data or training new models given new implementations.
However, most of these tools do stateless retraining, where the model is trained from scratch each time without leveraging efficient and adaptable ways to continually evaluate models.

Instead, CLaaS infrastructures are set up to do \emph{stateful} training. The latter is when you continue training your model on new data instead of retraining your model from scratch.
Therefore, instead of updating your models based on a fixed schedule, continually update your model whenever data distributions shift and the performance of the model decay. The training frequency is triggered by the drift detector.
Continual Learning strategies are applied to re-training the model in an efficient and adaptable manner. Furthermore, with the Continual Learning metrics, we can monitor model performance over time.

\subsection{Continual Learning featured in CLaaS}
The increasing demand for overcome classical machine learning process is leveraging the emergence of new solutions. 
For a company, CLaaS is a set of toolkit tools aimed to support the daily work of data scientists and data engineers in the machine learning development process. 
In particular, CLaaS provide features thought to manage the designing, building and managing of reproducible, testable ML-powered software.
It is specialized to guarantee Continuous Monitoring and Training using Continual Learning solutions.
In this subsection, it is described how CL methodologies are featured in the CLaaS paradigm with a brief review.

\textbf{Detection of data drift distribution}. 
In the classical ML, the data used in training follows a distribution supposed stationary.
Based on this assumption, the generalization of model can be limited by the distribution of data on which it was trained. 
Nevertheless, In a real-world application, we have a constant flow of information, where the distribution can change due to various external or internal factors. 
These changes in the input data distribution cause issues in previously trained models, mainly because their weights are not prepared to face the drift concerning the training data.
This problem creates the need to detect distribution shifts and update the model continually.

There are many types of concept drift that have been identified \cite{datadrift}, each of which can affect model performance. 
Different approaches were developed for the different kinds of shifts.
The CLaaS toolkit exploits both base and advanced approaches. 
In fact, the user can select among two main categories, supervised and unsupervised or build custom data drift approaches (for more detail of data drift solutions see Appendix in \cite{CLinPracticeAWS}).

\textbf{Continual Learning zoo algorithms}. 
There are many continual learning strategies in the literature developed for the neural network models. For a more in-depth overview, we refer the reader to the recent overviews in \cite{CLRobotics} and \cite{parisi2019continual}. The latter additionally exposes the bio-inspired aspects of existing continual approaches.
To roadmap in the CL strategies, it is useful to classify them into three main groups.
First, the Memory-based Continual Learning Methods, in which gather all methods that save raw samples as the memory of past tasks in episodic memory.
Second,  Architecture-based Continual Learning Methods, in which they use the model to overcome catastrophic forgetting and learn over time.
Typically, these classes of strategies exploit the dynamic neural network architecture changes that can be explicit or implicit.
Finally, Regularization-based Continual Learning Methods consist in modifying the update of weights when learning in order to keep the memory of previous knowledge. The literature on CL strategies propose also hybrid solutions \cite{CLRobotics}.

All these CL strategies allow the update of the model over time avoiding catastrophic forgetting and are in CLaaS.
In fact, the user of the CLasS can select among these different continuous learning approaches for different practical scenarios. 
Most of real-world use cases, Memory-based Methods can be performant at the cost of the major memory usage. 
If data preservation is a constraint, the user could select CL strategies that are Memory-free like Architectural or Regularization strategies.
Furthermore, in the edge computing context, the user can select efficient on-device CL strategies like \cite{hayes2022online} and \cite{pellegrini2020latent}.

\textbf{Continual Learning metrics}. 
A core feature not present in machine learning tools for the MLOps process is the use of the Continual Learning metric to detect model of the performance decay over time.
It is crucial to have a set of good evaluation metrics to monitor the model performance, i.e. the model forgets the past or does not learn new skills.
Detailed on CL evaluation protocols are well described in  \cite{CLRobotics} \cite{masana2020class} \cite{delange2021continual}.

Using Continual Learning strategies in CLaaS, it is possible to train models in a sequence of tasks and acquire new knowledge without forgetting what has been trained in the past. 
The evaluation criteria have to cover the whole aspect of the full (continual) learning problem. 
It is not enough to observe good final accuracy on an algorithm to know if it is transferable to a serving system.
We should also evaluate how fast it learns and forgets, if the algorithm is able to transfer knowledge from one task to another and if the algorithm is stable and efficient while learning and predict.

\section{Continual Brain: System Architecture and Implementation} 
\label{sec:system-DI} 

\begin{figure*}[ht!]
  \centering
   \includegraphics[width=0.95\linewidth]{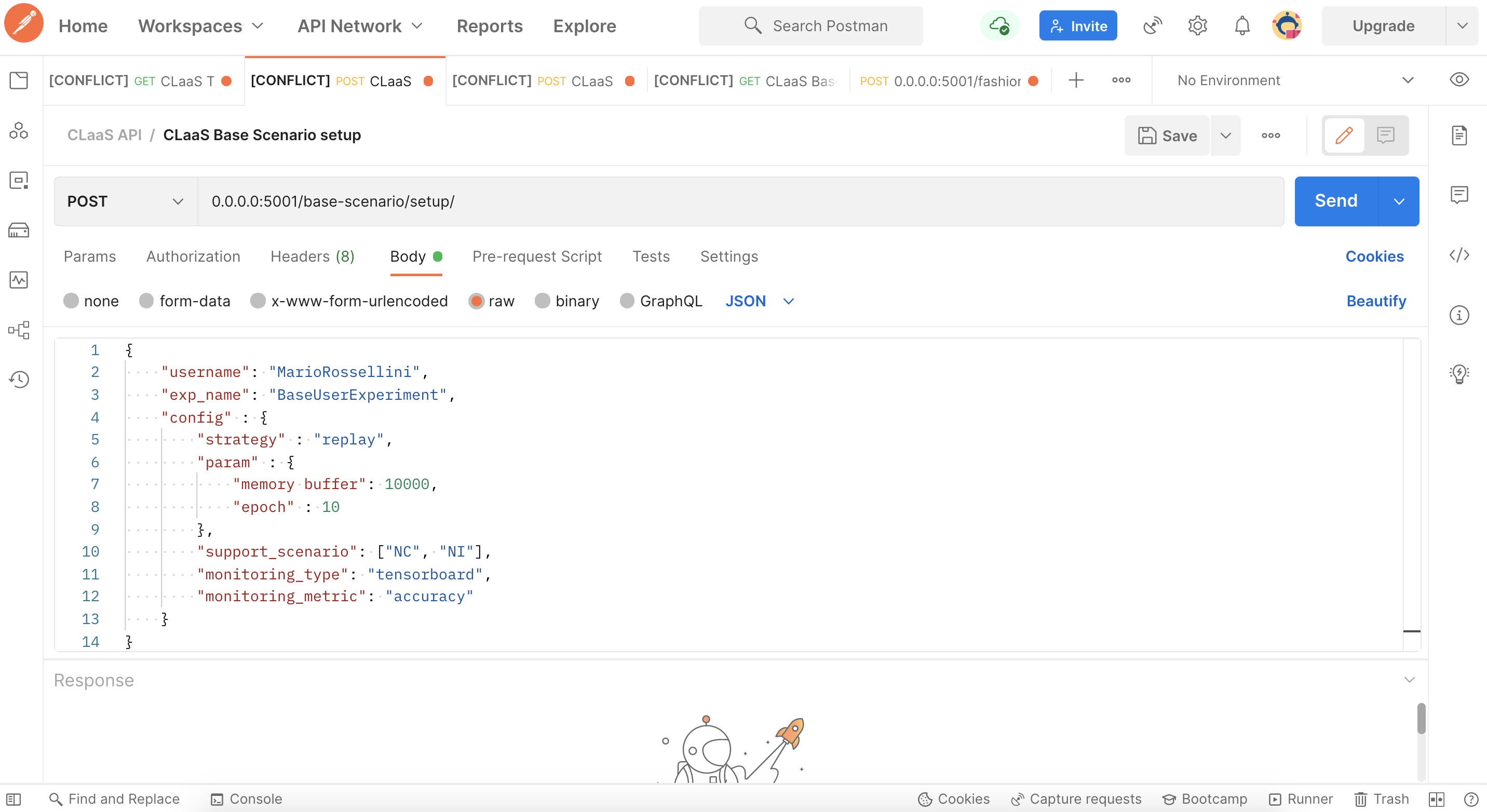}
   \caption{Example of user's request Restful API in JSON body (POSTMAN interface). Part of the configuration file to set up a typical experimental task in Continual Brain.}
   \label{fig: liquid-brain-config}
\end{figure*}

\subsection{Design Principles}
Continual Brain starts with a simple idea: extend the library Avalanche as a service model with a set of design guidelines: \emph{ modular and independent Building-Block view}, \emph{backend flexible and expandable}, \emph{portability}, \emph{ease-of-use}, \emph{efficiency and scalability}.
We believe that these principles are important first steps for reliable Continual Learning tools in real-world applications.

\textbf{Modular and independent Building-Block view} The main design principle for Continual Brain follows from the concept of modular building blocks, the idea of providing a set of services independent of each other and a baseline for further improvements.
This particular focus on module independence is maintained to guarantee the stand-alone individual module services. 
Moreover, the user can pick up a particular set of services and make use of a customize the others.
For instance, the user can select CL strategy services and customize services of CL metrics or CL models.

\textbf{Backend flexible and expandable}  Continual Brain is based on the principle of the API interface independent of the backend and easy to extend.
Therefore, it is possible for the maintainers to expand the backend and add more services API exposed to the user.
In this way, it allows also to employ of different environments, technologies and methods and guarantees flexibility of execution. 

\textbf{Portability} A critical design objective of Continual Brain is to allow experimental results to be seamlessly portable in both edge and cloud resources. 
As the first step, we have decided to allow data scientist and engineer to simply integrate their own research and code into the platform to speed up the development of original continual learning solutions in practical scenarios.

\textbf{Efficiency and scalability} These two designing principles are fundamental in modern DL research experiments.
Like current DL frameworks, we offer the end-user a seamless and transparent experience regardless of the use case or the hardware platform that the platform is run on.

\textbf{Ease-of-use} The last principle presented is the focus on simplicity.
All the Continual Learning services in Avalanche are given in a simple set of calls.
In particular, our efforts were focused on the design of an intuitive Application Programming Interface (API) and SW Development Kit (SDK) for data scientists and engineers.

\subsection{System Workflow}
The complete workflow of Continual Brain system can be split into two phases, the offline preparation phase and the online execution and monitoring phase. 

In the offline phase, researchers first leverage the built-in APIs or SDK of the CL zoo strategies to use or customize a CL model.
Meanwhile, engineers can prepare a configuration file to set up the trigger rule and the frequency of the retrain/fine-tuning according to our provided template.

In the online monitoring phase, the system follows the predefined rules in the configuration file to schedule model updating tasks. 
The main operation of monitoring model performance is based on CL metrics and it is executed in an automated manner with the user's configuration rules.

\subsection{Architecture}
The architecture of Continual Brain can be split into three main layers: \emph{Interface}, \emph{Middleware}, \emph{CL-Backend}.
Further internal components and frameworks model the overall structure, while the middle-layer endpoints are linked to the outermost components.

\textbf{Interface}, written in Flask python micro-framework, and helps the data scientist to interact with the CL-Backend. 
In particular, this layer manages communication with clients in RESTful APIs way or through an SDK (see Figure \ref{fig: liquid-brain-config}). 
This top layer interfaces with custom user code to run experiments or evaluate the performance with data tests. 

\textbf{Middleware}, a Python package that manages the monitoring and update stages, and controls model versioning. 
Moreover, this middle layer is set up as a proxy to access external storage services or for lower-level storage.
In this way, the logic of the lower-level storage is decoupled and flexible from the user storage preferences, making it possible also multiple storage services for the same user.
These middle-level components allow also services external to the application. The latter can be either integrated into the application or completely external. 
For example, data storage services and the database can be provided in the CL-backend or can be out cloud services.
Here, it is not used libraries but a custom codebase implementation.

\textbf{CL-Backend}, leverages the Avalanche CL library \cite{lomonaco2021avalanche} and PyTorch machine learning framework. 
The functionality of the Avalanche library in the Continual Brain service used are the CL zoo strategies, the CL models, the CL evaluation protocols and the easy set-up of CL scenarios (Table \ref{tab:CL-framework}).

Finally, Continual Brain was designed as a microservice architecture and implemented using the Docker Compose tool. 
The motivation is into the advantage of the architecture and tool for managing the application as a set of containers, giving the possibility to create multiple instances of the same service. The latter is important to guarantee the Continuous Training property in an MLOps manner.

\section{Empirical Evaluation} 
\label{sec:demo} 

Computer Vision (CV) and Deep Learning in robotics vision and fashion domain have become hot topics and received a great deal of attention in both academia and the industrial landscape \cite{CLRobotics} \cite{FashionCV}.
This section illustrates the advantage of CL as a model service in automating continuous model updating. In particular, we explore both object recognition in robotics and fashion real-world applications \cite{FashionCV2}, by discussing system efficiency when using a Continual Brain implementation.

\begin{figure}[t]
     \centering
     \includegraphics[width =\linewidth]{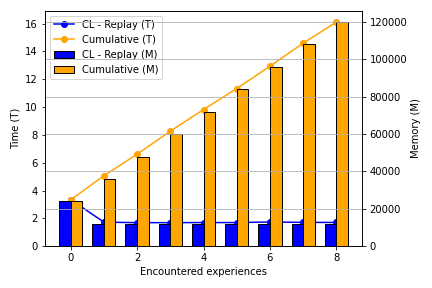}
   \caption{Cumulative and Replay CL-based strategy comparison (averaged on 3 runs) on CORe50 dataset: Computation in terms of time (min); memory in terms of the number of image patterns.}
    \label{fig:time_core50}
\end{figure}

\begin{figure}[t]
     \centering
     \includegraphics[width =\linewidth]{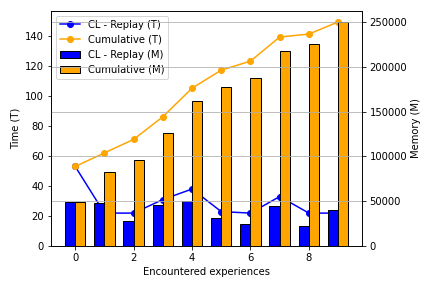}
   \caption{Cumulative and Replay CL-based strategy comparison (averaged on 3 runs) on DeepFashion-C dataset: Computation in terms of time (min); memory in terms of the number of image patterns.}
   \label{fig:time_df}
\end{figure}

\subsection{Experiment Setup} 
 All experiments are conducted with the following settings.
We use common  CL benchmark named CORe50 \cite{lomonaco2017core50} for Computer Vision object recognition application.
We also employ an instance of fashion analysis named categories prediction to evaluate both performance and efficiency. 
The latter is an enabling task for several fashion applications like visual search or visual recommendation \cite{FashionCV2}.
We use the DeepFashion-C dataset \cite{liu2016deepfashion} to build a new CL-benchmark in the fashion domain.
After splitting the dataset in train and test set, to simulate the CL process, we build a New Classes (e.g. NC, also known as \emph{Class-Incremental Learning}) scenario with 10 experiences.
Therefore, the first experiences contain 10 classes while the rest contain 4 classes.
The metric that is used to evaluate the clothing recognition models is the top-k accuracy.
The state-of-the-art methods to solve category and attribute prediction task in DeepFashion-C are summarized in  \cite{FashionCV2}.

To demonstrate the efficiency of the CL strategies in these practical domains, we compare a Cumulative approach with the Replay CL-based strategies \cite{parisi2019continual}. 
Cumulative is a stateless and time-consuming update process where all the data encountered through time are used to train the model in an offline manner.
We ran the experiments on a Multi-GPU NVIDIA-SMI server with 80-core Intel Xeon CPU E5-2698 v4 and 4 Tesla V100 GPUs 11.2 CUDA Version.

\subsection{Results}
\begin{figure}[t]
     \centering
     \includegraphics[width =\linewidth]{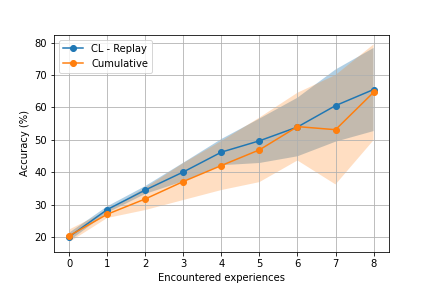}
   \caption{Cumulative and Replay CL-based strategy comparison (averaged on 3 runs) on CORe50 dataset: predictive performance in term of Accuracy using the entire test data for each experience.}
   \label{fig:performance_core50}
\end{figure}

\begin{figure}[t]
     \centering
     \includegraphics[width =\linewidth]{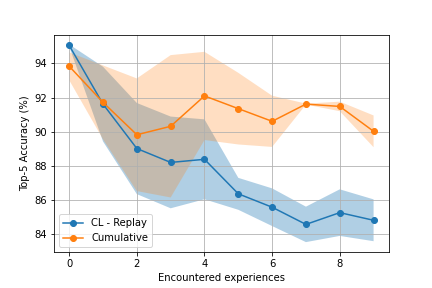}
   \caption{Cumulative and Replay CL-based strategy comparison (averaged on 3 runs) on DeepFashion-C dataset: predictive performance in term of Top-k accuracy with $k=5$ accumulating the test stream during the experiences.}
   \label{fig:performance_df}
\end{figure}

As Figures \ref{fig:performance_core50} and \ref{fig:performance_df} show, the strategies based on Continual Learning are very efficient at the cost of a small decay in predictive performance. As predicted, the Cumulative strategies achieve similar results of the upper bound but the cost grows with the number of encountered experiences.
In Figure \ref{fig:time_core50} we note that the time and memory saved by the CL approaches are up to a \textbf{$\times 6$} factor for a stream of 9 experiences. In Figure \ref{fig:time_df} this nears an \textbf{$\times 8$} factor in 10 experiences. The performance-efficiency trade-off is evident for the Replay CL strategy.
Finally, the Replay Continual Learning strategy has a stable training time in the growth of experiences. For this CL strategy the value $15000$ is used for the \emph{memory\_size} hyper-parameter in DeepFashion-C dataset. For CORe50 dataset the value of the \emph{memory\_size} are setting to $5000$.

These results demonstrate that the automation of the ML pipeline can be achieved  efficiently and in an adaptable manner with this CL-tool.
Therefore, in non-stationary production environments CL strategies might prove essential to attain efficient stateful training.

\section{Conclusion and Discussion} 
\label{sec:conlusion}
In this paper, we have described CLaaS, a novel software as a service and licensing model mainly based on a continual learning approach.
A first version, named Continual Brain, has been implemented by extending the functionalities of Avalanche as a set of micro-services.
We have demonstrated the usability and efficiency of the system with representative case studies in machine vision and fashion domains.

Nowadays,the industrial and company applications could help to direct new CL research directions toward more practical scenarios.
There are various practical fields that researchers and companies have investigated with Continual Learning.
The literature has proposed different CL real-world applications. 
For instance, in MLOps, surveillance videos, robotics and machine vision, clinical and medical sector, industry and edge computing.
Anyway, there are also many real-world and application-research directions unexplored with Continual Learning.

Prevalent scenarios of Continual Learning are Class Incremental Learning (CIL). They assume disjoint sets of classes as tasks but are less realistic in a real-world application.
Recent works \cite{RainbowBlurry} and \cite{i-Blurry} focus the attention to address a more realistic CL setup which occurs frequently in real world AI deployment scenarios. 
The request to practical CL setting seems also start from the companies that use CL for core business.

Finally, deploying CL to real-world applications is still quite challenging. 
The absence of CL tools in ML system workflow and the poor investigation of real-world applications are the main reasons. 
We believe the CLaaS paradigm can narrow the gap between research and its industrial application, and speed up companies innovative trend towards Continual Learning.

\newpage
\begin{footnotesize}

\bibliography{ref.bib}

\end{footnotesize}


\end{document}